\documentclass[letterpaper, 10 pt, conference]{ieeeconf}  
\usepackage{amsmath}
\usepackage{graphicx}
\usepackage{algorithm}
\usepackage{algpseudocode}
\usepackage{enumerate}
\usepackage{multirow}
\usepackage{subfig}

\IEEEoverridecommandlockouts                              

\title{\LARGE \bf
DyNaVLM: Zero-Shot Vision-Language Navigation System with Dynamic Viewpoints and Self-Refining Graph Memory
}

\author{Zihe Ji$^{1}$, Huangxuan Lin$^{2}$ and Yue Gao$^{3 \dag}$
\thanks{This work was supported by the National Natural Science Foundation of China (Grant No. 62373242 and No.92248303),Shanghai Municipal Science and Technology Major Project (Grant No. 2021SHZDZX0102).
} 
\thanks{$^{1}$Zihe Ji is with SJTU Paris Elite Institute of Technology, Shanghai Jiao Tong University, Shanghai, P.R. China. {\tt\small william\_ji@sjtu.edu.cn}}%
\thanks{$^{2}$Huangxuan Lin is with Department of Automation, Shanghai Jiao Tong University, Shanghai, P.R. China. {\tt\small b3mylq@sjtu.edu.cn}}.
\thanks{$^{3}$Yue Gao is with MoE Key Lab of Artificial Intelligence and AI Institute, Shanghai Jiao Tong University, P.R. China, {\tt\small yuegao@sjtu.edu.cn}. $^\dag$Yue Gao is the corresponding author.}%
}

\begin{document}

\maketitle
\thispagestyle{empty}
\pagestyle{empty}

\begin{abstract}
    We present DyNaVLM, an end-to-end vision-language navigation framework using Vision-Language Models (VLM). In contrast to prior methods constrained by fixed angular or distance intervals, our system empowers agents to freely select navigation targets via visual-language reasoning. At its core lies a self-refining graph memory that 1) stores object locations as executable topological relations, 2) enables cross-robot memory sharing through distributed graph updates, and 3) enhances VLM's decision-making via retrieval augmentation. Operating without task-specific training or fine-tuning, DyNaVLM demonstrates high performance on GOAT and ObjectNav benchmarks. Real-world tests further validate its robustness and generalization. The system's three innovations—dynamic action space formulation, collaborative graph memory, and training-free deployment—establish a new paradigm for scalable embodied robot, bridging the gap between discrete VLN tasks and continuous real-world navigation.

\end{abstract}

\section{Introduction}

Navigation is a foundational capability for autonomous agents, requiring the integration of spatial reasoning, real-time decision-making, and adaptability to dynamic environments. While humans navigate seemingly effortlessly through complex spaces, replicating this ability in artificial systems remains a formidable challenge. Traditional approaches often decompose the problem into modular components—such as perception, reflection, planning, and control—to leverage specialized algorithms for each subtask.\cite{c1}\cite{c2} However, such systems struggle with generalization, scalability, and real-world deployment due to their reliance on task-specific engineering and rigid pipelines. Recent advances in VLM \cite{c3}\cite{c4} offer promising alternatives by unifying perception and reasoning within a single framework, yet their application to embodied navigation remains constrained by limitations in spatial granularity and contextual reasoning.

In this work, we demonstrate that VLM can function as a zero-shot, end-to-end navigation policy without requiring fine-tuning or prior exposure to navigation-specific data. Our approach is built on three key innovations. First, we introduce a dynamic action space formulation that replaces fixed motion primitives with free-form target selection through visual-language reasoning. Second, we develop a collaborative graph memory mechanism, inspired by retrieval-augmented generation (RAG), which enables persistent spatial knowledge representation and cross-robot memory sharing. Third, our method allows for training-free deployment, achieving high performance without task-specific fine-tuning.  

\begin{figure}[t]
\centerline{\includegraphics[width=0.9\columnwidth]{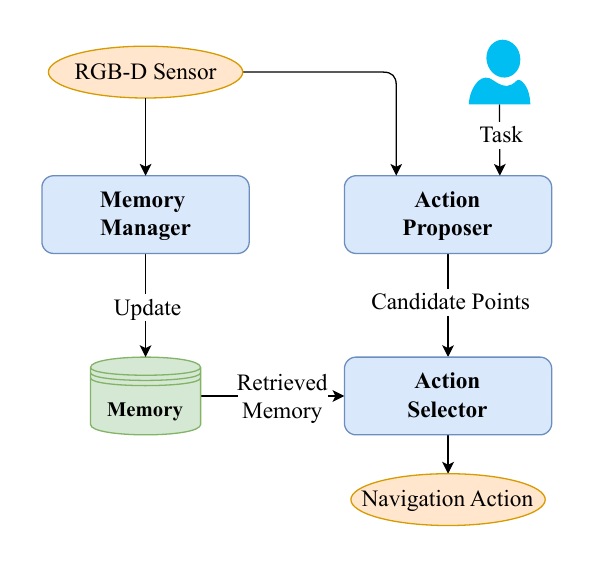}}
\caption{Overview of the proposed framework, which consists of three main components: (1) Memory Management, where the RGB-D sensor captures raw data, and the memory manager updates the memory bank; (2) Action Proposer, where the action proposer generates candidate navigation points based on perception; and (3) Decision Making, where the action selector retrieves relevant memory information and determines the final navigation action. Blue boxes represent processing modules, the green cylinder denotes memory storage, and red shapes indicate data flow.}
\label{fig1}
\end{figure}

At the core of our system is a self-refining graph memory architecture that encodes object relationships as executable topological maps, supports distributed memory updates across robots, and enhances VLM decision-making through context-aware memory retrieval. This architecture bridges the sim-to-real gap by allowing agents to dynamically adjust navigation targets based on safety constraints, such as obstacle avoidance and adherence to warning signs, while maintaining human-interpretable spatial representations. Fig.~\ref{fig1} 

Our contributions are threefold. First, we introduce a novel graph memory architecture that encodes environments as executable topological maps, enabling persistent spatial reasoning and facilitating distributed knowledge sharing across robotic agents. Second, we present DyNaVLM, a novel VLM-based navigation framework capable of achieving human-level target selection flexibility, effectively eliminating the constraints of fixed action spaces. Third, it provides comprehensive validation demonstrating high-performance in both simulated benchmarks and real-world deployment, showcasing the framework's robustness for practical applications without requiring task-specific training.

\section{Related work}

\subsection{Vision-Language Models for Navigation}
One of the most common vision-language navigation (VLN) policies is training end-to-end models from scratch using offline datasets\cite{c5}\cite{c6}\cite{c7}\cite{c8}\cite{c9}, but face fundamental limitations in generalization due to the prohibitive costs of collecting large-scale navigation data. These methods struggle to adapt to novel object configurations or unseen environments.An alternative line of work fine-tunes pre-trained vision-language models with robot-specific data \cite{c10}\cite{c11}\cite{c12}\cite{c13}, aiming to preserve semantic understanding while adapting to navigation tasks. However, this risks catastrophic forgetting of the original model's capabilities, revealing the fragility of such approaches. The cost of fine-tuning is also non-negligible, requiring extensive data collection and annotation efforts.

Recent efforts have shifted toward zero-shot prompting of VLMs without fine-tuning, but these methods impose restrictive action spaces such as fixed angular increments and cardinal directions \cite{c7}\cite{c14}\cite{c15} or directions ("north," "south", "east," "west") \cite{c2} that limit motion flexibility. This fixed-distance and fixed-direction action space either oversimplifies the situation or relies on extremely small distances or angles, causing the robot to make repeated small steps, which leads to an unacceptable time cost. In any case, such action spaces remain merely an idealized abstraction.

Visual prompting methods have emerged to enhance spatial reasoning in VLMs. Building on annotation techniques like Set-of-Mark \cite{c16}, navigation-specific approaches overlay markers on input images to ground language instructions in visual features \cite{c17}. PIVOT \cite{c18} and VLMnav \cite{c19} represent this category, generating action proposals through arrow prompts. While these methods improve navigation performance, they still rely on fixed action spaces and lack consideration for hard constraints (walkable areas) and soft constraints (such as warning signs or task requirements).

\subsection{Memory for Navigation}
Effective navigation in mobile robotics requires robust environmental memory systems that integrate sensory perception with spatial modeling. Traditional approaches build upon Simultaneous Localization and Mapping (SLAM) \cite{c20}, which constructs geometric maps through sensor fusion and localization. While SLAM enables basic obstacle avoidance and path planning, its purely geometric representations lack semantic context, limiting applicability in human-centric environments and cannot be directly used for VLM.

This semantic gap has spurred development of augmented memory systems that layer object-level understanding onto spatial maps. Current solutions incorporate semantic information through two primary modalities: direct human interaction via dialogue systems \cite{c21}, or automated annotation using machine learning models. The latter encompasses hierarchical 3D scene graphs encoding room connectivity\cite{c22}, vision-language models generating textual scene descriptions \cite{c23}, and region classifiers labeling navigable spaces in RGB-D streams \cite{c24}. Despite these advances, implementation barriers persist—particularly dependence on specialized sensors and compute-intensive hardware—which restrict real-world deployment scalability, especially unable to annotate the environment for a robot in real time. Existing methods thus face a fundamental tradeoff: geometric precision and semantic richness remain inversely correlated with practical feasibility across diverse robotic platforms. 

Recent works like SayPlan \cite{c21} and Text2Map \cite{c25} demonstrate graph-based mapping approaches that encode spatial relationships through hierarchical structures. While we similarly adopt graph representations for location encoding, these methods rely on specialized prior knowledge—SayPlan requires pre-constructed 3D scene graphs (3DSG) of large environments, while Text2Map depends on human-provided navigation instructions for graph initialization. In contrast, our framework integrates memory construction as an inherent component of exploration, enabling VLMs to dynamically build graph-structured memory without prior environmental knowledge. This dual approach ensures robustness through self-supervised map generation while eliminating dependence on human annotations. 

\section{Method}

\subsection{Problem Formulation} 
We introduce DyNaVLM, a navigation system that processes a goal \( g \) (expressed in either natural language or an image), an RGB-D observation \( I_t \) at every step, and the robot's pose \( \xi_t \) to determine an appropriate action \( a_t \in A_t\). The agent maintains a memory function $\mathcal{M}_t = \Phi(\{I_k, \xi_k\}_{k=0}^t)$ that extracts spatio features from historical observations. The action space is parameterized in polar coordinates $a_t = (r_t, \theta_t)$, comprising a yaw-axis rotation and a forward displacement in the robot’s local frame. In summary, at every step $t$, our policy $\pi$ computes an action $a_t = \pi(s_t, \mathcal{M}_t)$, where $s_t = <I_t, \xi_t, g>$.

\subsection{Memory manager}\label{sec:memory_manager}

\begin{figure*}[t]
    \centering
    \subfloat{
        \includegraphics[width=0.45\linewidth]{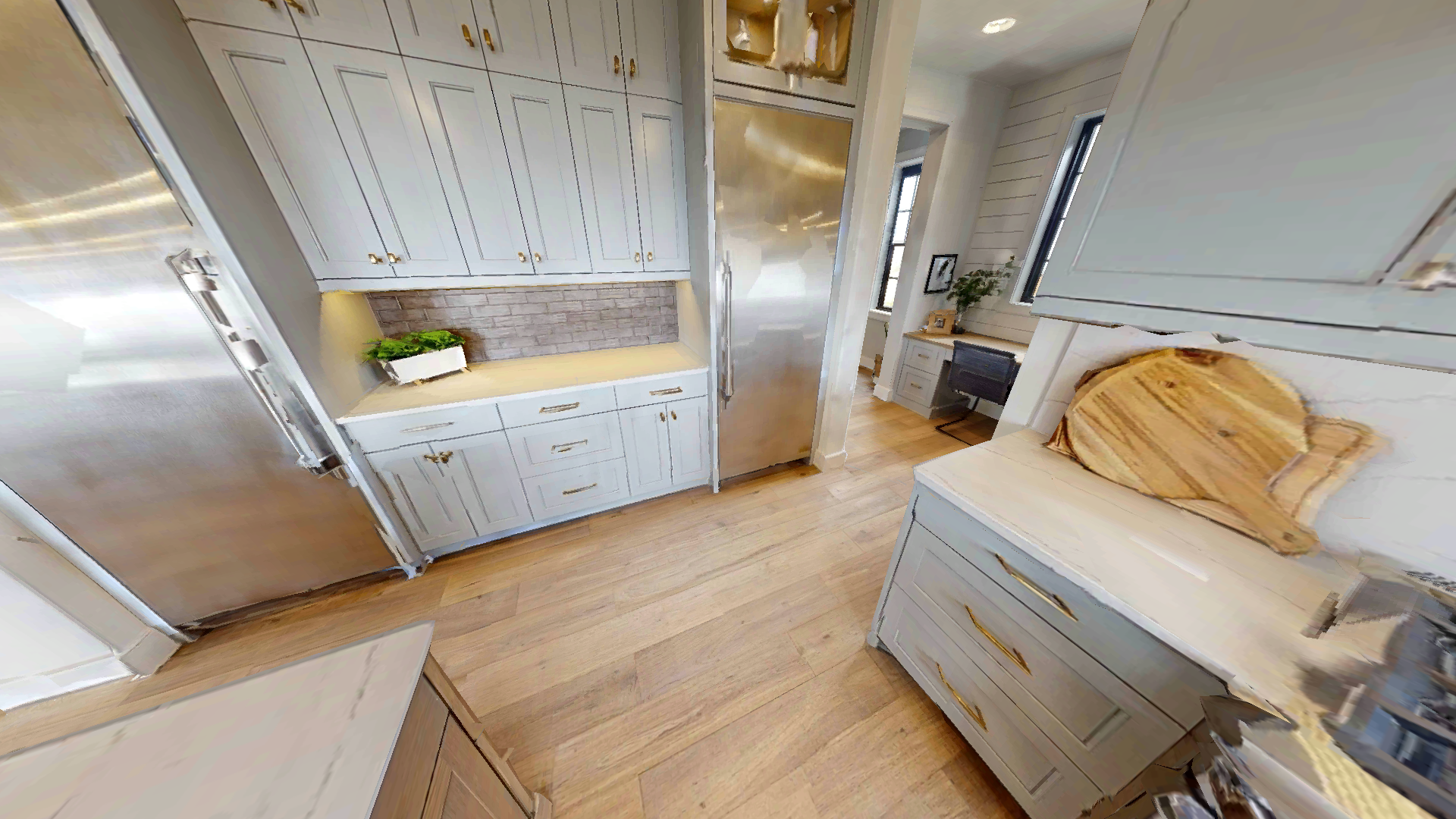}
        \label{fig:rgbd_input}
    }
    \hfill
    \subfloat{
        \includegraphics[width=0.52\linewidth]{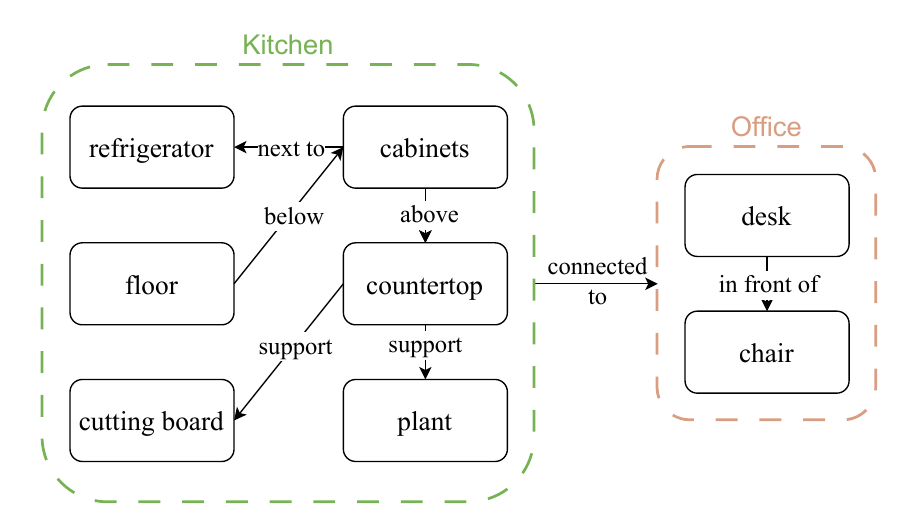}
        \label{fig:vlm_output}
    }
    \caption{Illustration of the input image and processing results of the Memory Manager. (a) The raw RGB data captured by the sensor. (b) The structured information extracted by the vision-language model, where each node in the structure represents an object along with its attributes, such as \texttt{"refrigerator": ["stainless steel"]}, \texttt{"cabinets": ["white"]}, \texttt{"cutting board": ["wooden"]}, \texttt{"plant": ["green", "potted"]}, etc. The edges encode spatial relationships between objects, such as \textit{next to}, \textit{below}, \textit{above}, \textit{in}, \textit{in front of} and so on. This structured memory captures implicit hierarchical relationships—e.g., \texttt{"kitchen"} and \texttt{"office"} can be considered as parent nodes—enabling a more abstract and structured map representation.}
    \label{fig:memory}
\end{figure*}

The Memory manager serves as the central middleware coordinating the interaction between the VLM and our graph-structured memory system. As illustrated in Fig. \ref{fig:memory}, this component maintains a dynamic knowledge graph (Memory Bank) that captures spatial relationships and semantic object information within the environment. 

The graph structure of memory $G = (V, E)$ implements a hybrid representation where vertices $v_i \in V$ encode object instances with attributes \texttt{\{name, location, visual\_features\}}, while edges $e_{ij} \in E$ model spatial adjacency and navigation accessibility. Upon receiving real-time perceptual input from RGB-D sensors, the manager performs two critical functions: (1) node embedding updates using VLM-generated object descriptors, (2) edge addition, where the VLM determines spatial relationships between nodes. The VLM can update the memory through functions 
\begin{itemize}
    \item \texttt{add\_node(name, attributes)}: Create or update a node in the memory.
    \item \texttt{add\_edge(start, target, relation)}: Add a directed edge between two nodes.
\end{itemize}

The manager exposes a structured query interface to the LLM controller through function templates including:
\begin{itemize}
    \item \texttt{spatial\_query(semantic\_filter)}: Retrieves neighborhood subgraphs based on semantic constraints.
    \item \texttt{path\_inference(start, target)}: Generates potential navigation trajectories.
\end{itemize}

This architecture facilitates efficient context propagation between the VLM's visual grounding capabilities and the LLM's symbolic reasoning, which is particularly crucial for handling ambiguous object references (e.g., "find the charging station near seating areas"). Additionally, this structured memory can be easily converted back into natural language to describe objects and spatial relationships within the memory, serving as input for the VLM. Moreover, it enables seamless memory transfer and preservation.

\begin{figure*}[t]
    \centering
    \subfloat{
        \includegraphics[width=0.23\linewidth]{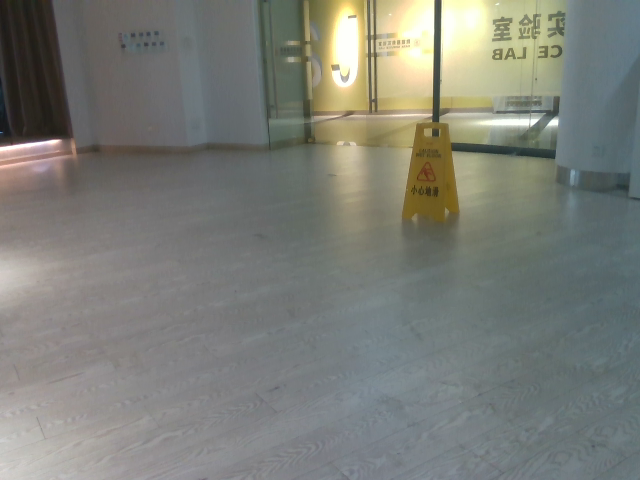}
        \label{rgb}
    }
    \hfill
    \subfloat{
        \includegraphics[width=0.23\linewidth]{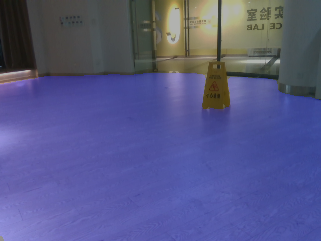}
        \label{gnd}
    }
    \hfill
    \subfloat{
        \includegraphics[width=0.23\linewidth]{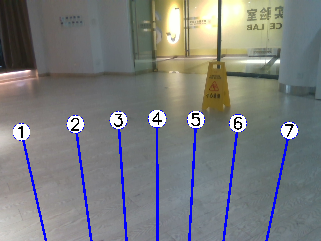}
        \label{init}
    }
    \hfill
    \subfloat{
        \includegraphics[width=0.23\linewidth]{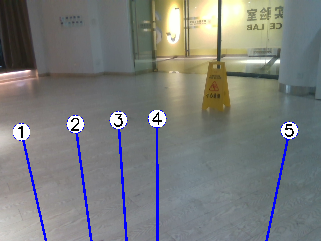}
        \label{final}
    }
    \hfill
    \caption{Illustration of the Action Proposer process.  (a) The input RGB image, showing a "Caution: Wet Floor" sign on the floor.  (b) The navigable ground region is extracted using depth data, with the ground highlighted in blue.  (c) Initial candidate points \( A_{\text{initial}} \) are generated, but points 5 and 6 are too close to the caution sign.  (d) The VLM refines the candidate points by incorporating the natural language constraint ("avoid the caution sign"), removing points 5 and 6 to ensure safe navigation.}
    \label{fig:action_proposer}
\end{figure*}

\subsection{Action Proposer}\label{sec:action_proposer}
Enabling a VLM to directly output navigation coordinates from an image is inherently challenging, as spatial reasoning is not explicitly encoded in standard language model training objectives. To bridge this gap without introducing task-specific training, we propose a candidate-based discretization strategy that simplifies the continuous search space into a finite set of uniformly sampled points. These candidate points are generated without prior environmental assumptions, ensuring minimal human bias and maximizing adaptability to novel scenes. The action proposal process involves two stages:

\subsubsection{Spatial Sampling}
Given RGB-D data $\mathbf{I}, \mathbf{D}$, we define the navigable boundary as a continuous set of maximum reachable ground points in polar coordinates:

\begin{equation}
\mathcal{G}_{\text{line}} = \left\{(r, \theta) \mid r = \max_{\text{ground}} d(\theta)\right\}
\end{equation}

where $d(\theta)$ represents the depth measurement at angular coordinate $\theta$.

With uniform angular sampling, we generate a set of candidate points $A = \{(r, \theta)\}$ along the navigable boundary. To ensure the visual spacing between actions is enough for the VLM to discern, the points are filted by a minimum angular distance $\theta_{\delta}$. To encourage exploration, the near points are removed in priority to the far points. A coefficient $\alpha < 1$ is multiplied to the distance to ensure a safe margin.

\begin{equation}
A_{\text{initial}} = \left\{(\alpha r, \theta)  \bigg| (r, \theta) \in A, \left|\theta_i-\theta_j\right| \geq \theta_{\delta}, \forall i,j \right\}
\end{equation}

\subsubsection{Safety-Aware Filtering}
VLM is used to further filter and sdjust the candidate points. The coordinates of the candidate points are concatenated with the natural language constraints $\mathcal{C}$ to form the input to the VLM, and points are also annoted on the image $\hat{\mathbf{I}}$.
To ensure real-world applicability, candidate points undergo rigorous safety checks. Our framework dynamically eliminates infeasible waypoints (e.g., near obstacles, on non-traversable surfaces) caused by ground recognition errors (see Fig.~\ref{fig:action_proposer}). Natural language-defined constraints (e.g., "stay away from the oven") can also be involved. 

Moreover, for surviving candidates, the VLM performs geometric adjustments based on visual context. By analyzing scene semantics (e.g., objects, signage), the model fine-tunes point positions or rotating headings to better align with task objectives. This refinement process leverages the VLM's commonsense understanding of objects, enabling precise navigation without relying on domain-specific object detectors or geometric priors.

\begin{equation}
A_{\text{final}} = \text{VLM}\left(\hat{\mathbf{I}}, \mathcal{C}, A_\text{initial}\right)
\end{equation}

This dual-stage approach exemplifies our framework's generalizability: it avoids hardcoding environmental assumptions while maintaining robustness through the VLM's inherent world knowledge. By decoupling spatial sampling from semantic reasoning, we preserve the model's zero-shot capabilities while ensuring safety-critical constraints are systematically enforced.

\subsection{Action Selector}
The Action Selector synthesizes geometric candidates, perceptual context, and retrieved memory to produce the final navigation action. 

\subsubsection{Multimodal Prompt Engineering}
Given candidate points $A_{\text{final}} = \{(r_i, \theta_i)\}$ from \ref{sec:action_proposer} and retrieved memory $\mathcal{M}$ from \ref{sec:memory_manager}, we construct a multimodal prompt combining:
\begin{equation}
    \mathcal{P} = [\hat{I} \oplus T \oplus \mathcal{M}]
\end{equation}
where $\hat{I}$ denotes the annotated RGB image, $T$ the task-specific template, and $\mathcal{M}$ the structured memory entries retrieved through the memory retrieval function.

\subsubsection{Confidence-Based Execution}
Using chain-of-thought reasoning, the VLM evaluates candidate actions $a_i \in A_{\text{final}}$ based on the multimodal prompt $\mathcal{P}$ to generate a confidence score $s_i$ for each action. To determine whether terminate the navigation, a stop action is checked $s_{\text{stop}} = \mathcal{V}(I, T_{\text{stop}})$. The original image $I$ is used to verify the stop action. The final action $a^*$ is selected:
\begin{equation}
    a^* = \begin{cases}
        \text{stop}, & \text{if } s_{\text{stop}} > \tau_{\text{stop}} \text{ for consecutive 2 steps} \\
        \underset{a_i \in A_{\text{final}}}{\arg\max}\ s_i, & \text{otherwise }
    \end{cases}
\end{equation}

This architecture enables zero-shot adaptation to novel environments while maintaining safety constraints through the memory-guided verification process.

\section{Simulation Evaluation}
\begin{figure*}[t]
    \centering
    \includegraphics[width=\linewidth]{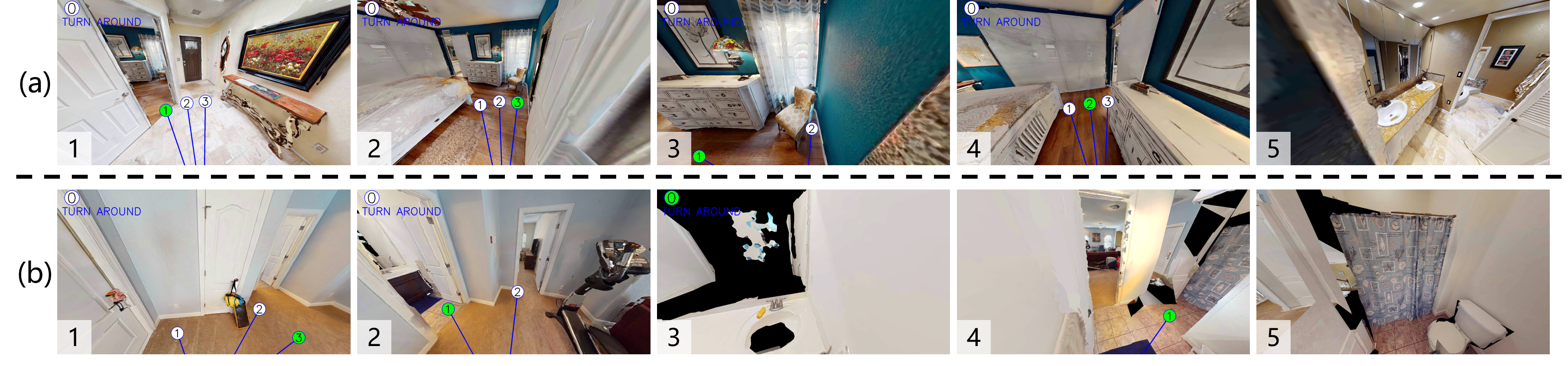}
    \caption{Examples from the simulation environment. (a) A task from GOAT-Bench with the goal description: "Mirror in the bathroom. Start by locating the bathroom counter and the sink. The mirror should be located near the sink. Look for a reflective surface attached to the wall. It is a bathroom mirror." (b) A task from ObjectNav, where the goal is to navigate to the toilet.}
    \label{fig:simulation}
\end{figure*}
To assess and compare our approach with the prior ones, performances are compared on two embodied navigation benchmarks, ObjectNav \cite{c26} and GOAT-Bench \cite{c27}, with the HM3D dataset \cite{c28}. Further, ablation studies are conducted to evaluate the effectiveness of memory in our framework. 

\subsection{Experiment Setups}
We adopt the same setup as VLMnav~\cite{c19}. The agent has a cylindrical body with a radius of 0.17m and a height of 1.5m. It is equipped with an egocentric RGB-D sensor with a resolution of $(1080, 1920)$ and a horizontal field-of-view  of $131^\circ$. We use Gemini 2.0 Flash-Lite as the vision-language model.

\subsection{Metrics} \label{simulation metrics}
Following ~\cite{c2, c19, c27}, we evaluate performance using:
\begin{itemize}
    \item \textbf{Success Rate (SR)}: The fraction of episodes completed successfully.
    \begin{equation}
        SR = \frac{N_{\text{success}}}{N_{\text{total}}}
    \end{equation}
    where \( N_{\text{success}} \) is the number of successfully completed episodes, and \( N_{\text{total}} \) is the total number of episodes.
    
    \item \textbf{Success weighted by Path Length (SPL)}: A path efficiency measure penalizing unnecessary movement.
    \begin{equation}
        SPL = \frac{1}{N} \sum_{i=1}^{N} S_i \frac{l_i}{\max(p_i, l_i)}
    \end{equation}
    where \( S_i \) is a binary success indicator for episode \( i \), \( l_i \) is the shortest path length from start to goal in episode \( i \), and \( p_i \) is the actual path length taken by the agent.
\end{itemize}

\subsection{Results}
 
\subsubsection{ObjectNav}
The Object-Goal Navigation (ObjectNav) benchmark, which evaluates an agent’s ability to navigate to an object specified by its category label (Sofa, Toilet, TV, Plant, Chair, Bed) in an unexplored environment. The agent starts at a random location and must explore efficiently to approach the target. The threshold of success is 0.3m.

We compare our approach against PIVOT~\cite{c5} and VLMnav~\cite{c19} , the most relevant baselines using VLM. We also evaluate an ablated variant, Ours w/o Memory, which removes the Memory Manager modules.

\begin{table}[h]
\caption{ObjectNav Results}
\label{table_ObjectNav}
\begin{center}
\begin{tabular}{|c||c|c|}
\hline
Method & SR (\%) & SPL \\
\hline
Ours & \textbf{45.0} & \textbf{0.232} \\
w/o Memory & 43.1 & 0.187 \\
VLMnav~\cite{c19} & 42.8 & 0.218 \\
PIVOT~\cite{c5}& 22.7 & 0.108 \\
\hline
\end{tabular}
\end{center}
\end{table}

Table~\ref{table_ObjectNav} shows that our approach achieves the best performance on ObjectNav, with a 45.0\% success rate (SR) and an SPL of 0.232, outperforming the previous VLM frameworks. The ablated variant Ours w/o Memory, which removes the memory module, sees a performance drop on the SPL, demonstrating the effectiveness of our memory design in improving both success rate and path efficiency. Moreover, both Ours and Ours w/o Memory outperform PIVOT~\cite{c5}. This suggests that restricting the sampling of navigation points to the ground plane is a crucial strategy, allowing for more efficient exploration and goal localization.

\subsubsection{GOAT-Bench}
GO to AnyThing (GOAT) benchmark~\cite{c27} is a challenging benchmark for embodied AI navigation, requiring agents to follow targets specified through three modalities: (i) object names, (ii) object images, and (iii) detailed text descriptions (as shown in Fig.~\ref{fig:simulation}). Each episode consists of 5–10 sub-tasks, testing the agent's adaptability to different goal representations. GOAT-Bench evaluates methods on multimodal navigation, robustness to noisy goals, and memory usage in lifelong scenarios.

Besides comparing to VLMnav and PIVOT, We also evaluate our method against two state-of-the-art specialized approaches: (i) SenseAct-NN [17], a reinforcement learning-based policy that employs learned submodules for different navigation skills, and (ii) Modular GOAT [20], a system that constructs a semantic memory map and utilizes a low-level policy for object navigation.

Unlike SenseAct-NN and Modular GOAT, our approach is zero-shot and requires no task-specific training. We do not depend on low-level policies or external object detection/segmentation modules. Instead, our method leverages a generalizable memory module, allowing it to achieve competitive performance with these specialized approaches while maintaining flexibility across different goal modalities.

\begin{table}[h]
\caption{GOAT-Bench Results}
\label{table_GOAT}
\begin{center}
\begin{tabular}{|c||c|c|}
\hline
Method & SR (\%) & SPL \\
\hline
SenseAct-NN Skill Chain~\cite{c27} & \textbf{29.5} & \textbf{0.113} \\
Ours & \textbf{25.5} & 0.102 \\
Modular GOAT~\cite{c29} & 24.9 & \textbf{0.172} \\
VLMnav~\cite{c19} & 20.1 & 0.096 \\
PIVOT~\cite{c5}& 10.2 & 0.055 \\
\hline
\end{tabular}
\end{center}
\end{table}

As shown in Table~\ref{table_GOAT}, our method achieves strong performance on the GOAT-Bench benchmark. Among all VLM-based approaches, our method achieves the best performance, demonstrating that a well-designed memory module can significantly enhance the generalization capability of VLMs for embodied navigation. While SenseAct-NN~\cite{c27} attains a 15\% higher performance, our method achieves \textbf{25.5\%} SR and 0.102 SPL, demonstrating competitive results despite being a zero-shot approach without task-specific training. 

Compared to Modular GOAT~\cite{c29}, our method achieves a higher SR, while Modular GOAT attains a higher SPL (0.172 vs. 0.102). This suggests that the object recognition model, which provides precise object locations for navigation, allows the agent to take more direct and efficient paths to the target, improving its SPL. In contrast, our approach relies solely on a vision-language model (VLM) with memory to reason about spatial relationships, achieving competitive success rates without explicit object localization. 

Moreover, our method outperforms VLMnav~\cite{c19} by 27\% in SR, demonstrating the substantial benefit of our memory module in longer and more complex navigation tasks. Since GOAT-Bench includes multiple sub-tasks in navigation episodes and noisy goal specifications, memory plays a crucial role in helping the VLM track previously visited locations, recognize partially observed objects, and efficiently plan future movements. This further supports our hypothesis that memory is especially valuable in lifelong learning settings and real-world embodied AI scenarios, where agents must retain and utilize past knowledge over extended durations.

\section{Real-world Evaluation}
We condected experiments under various real-world tasks to verify the effectiveness of the proposed method. 

\begin{figure*}[t]
    \centering
    \includegraphics[width=0.8\linewidth]{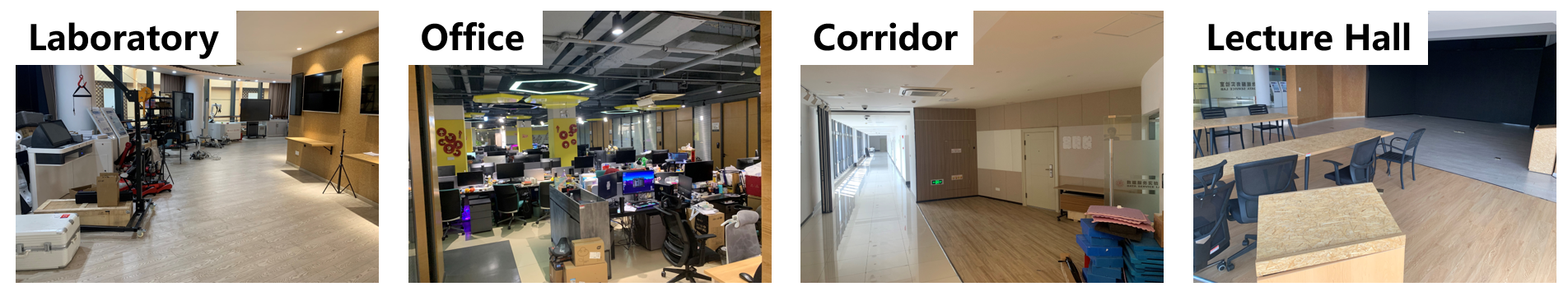}
    \caption{Overview of the main scenarios that constitute the real-world evaluation environment.}
    \label{fig:real-world_scene}
\end{figure*}

\begin{figure*}[t]
    \centering
    \includegraphics[width=\linewidth]{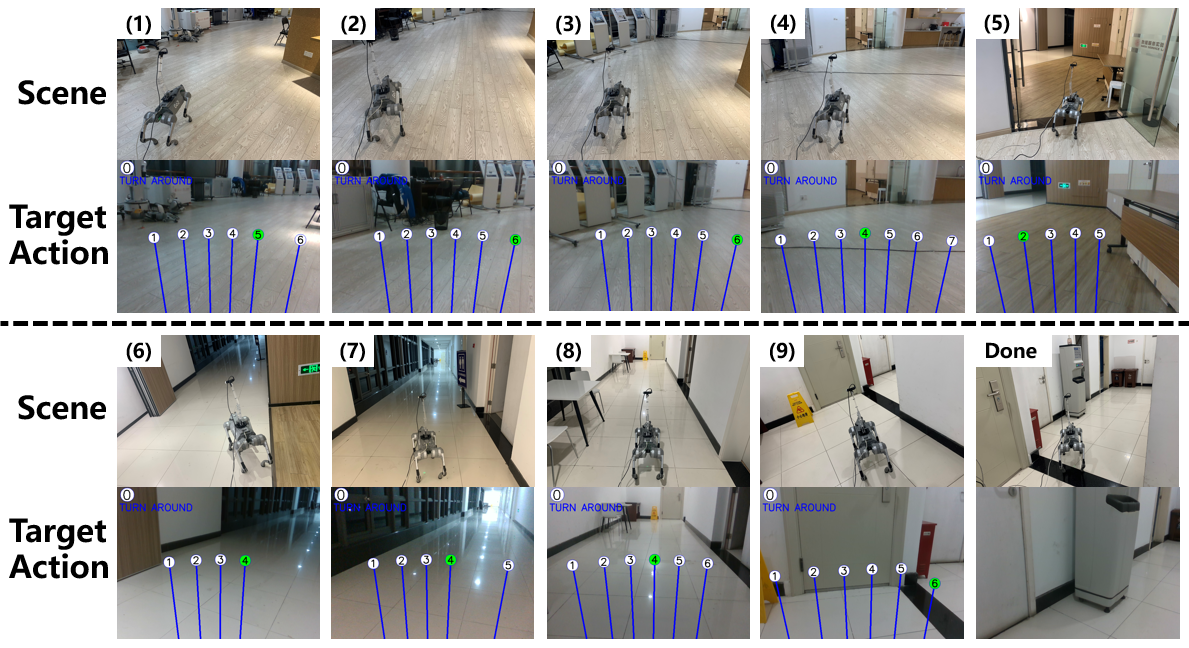}
    \caption{Key frames that describe one of the real-world evaluation task, where the agent is required to navigate to the nearest water dispenser.}
    \label{fig:real-world_exp}
\end{figure*}

\subsection{Experiment Setups}
The proposed method was deployed on a Unitree Go2 quadrupedal robot. A RealSense D435 camera is used to acquire aligned RGB-D images, which is mounted at a downward angle of 10 degrees and a height of 85cm (related to the ground). A Unitree L1 LiDAR located beneath the robot’s head is used to provide local odometry for controlling the robot’s movement towards the target point. The experiments are deployed on a large flat floor containing open-plan offices of different sizes, meeting rooms, lecture hall and laboratory space. The overview of the scenarios are shown in Fig. \ref{fig:real-world_scene}.

In each task, the agent is prompted to find 1 or 2 nearest objects. We employed FastSAM~\cite{c30} for the segmentation of the ground in the RGB image. For the determined target point, we project its position into the robot’s base frame using the aligned depth image. Coordinate transformation between the robot and the target position is updated from the local odometry to command the robot’s movement using the default motion controller of Go2. To reserve space for the movement, the L1 LiDAR is also used to detect surrounding obstacles. When an obstacle is detected within a cylindrical area whose diameter exceeds the length of robot by 15cm, the robot will first move a certain distance in the opposite direction of the obstacle, and then proceed towards the target.

\subsection{Metrics}
The performance in real-world tasks are evaluated using the Success Rate (SR) metric mentioned in Sec. \ref{simulation metrics} and the average cumulative distance (ACD) traversed across all succeeded tasks. 

The success of real-world tasks is defined as the distance between the target object and the robot being less than 5m, with the target appearing within the last frame of camera. For the first target, the robot is allowed a maximum travel distance of 200m, which is approximately twice the distance required to traverse the entire experimental scenario. For the second target (if given), the robot is permitted an additional maximum travel distance of 100m.

\subsection{Results}
We conducted five tasks in total, with three tasks requiring the agent to locate single targets: a water dispenser, a printer, and a trash can. In the remaining two tasks the agent were sequentially required to find two targets, specifically first the printer and then the trash can, and first the water dispenser and then the printer. To enhance the diversity of the tasks, we conducted three separate experiments for each task, initiating from distinct locations.
  
We compared the proposed method with the VLMnav baseline~\cite{c19}, ensuring fairness by starting from the same location and direction for the same experiment. The experimental results are shown in Table \ref{tab:real-world_result}. The superscript “1” denotes the target as the first instructional objective, while the superscript “2” signifies the target as the second instructional objective. The result indicates that our method notably outperforms the baseline in terms of both success rate and travel distance, with an particularly pronounced improvement in tasks requiring the identification of the second target. 

\begin{table}[h]
\caption{Result of the real-world evaluation}
\label{tab:real-world_result}
\begin{center}
\begin{tabular}{|c||cc|cc|cc|}
\hline
\multirow{2}{*}{} & \multicolumn{2}{c|}{Printer$^1$}     & \multicolumn{2}{c|}{Water dispenser$^1$} & \multicolumn{2}{c|}{Trash can$^1$}           \\ \cline{2-7} 
                  & \multicolumn{1}{c|}{SR. $\uparrow$} & ACD. $\downarrow$ & \multicolumn{1}{c|}{SR. $\uparrow$}   & ACD. $\downarrow$   & \multicolumn{1}{c|}{SR. $\uparrow$}    & ACD. $\downarrow$    \\ \hline
Ours              & \multicolumn{1}{c|}{2/6}   &    89m    & \multicolumn{1}{c|}{3/6}     &   65m       & \multicolumn{1}{c|}{2/3}      &     29m      \\ \hline
VLMnav             & \multicolumn{1}{c|}{1/6}   &    23m    & \multicolumn{1}{c|}{1/6}     &    44m     & \multicolumn{1}{c|}{2/3}      &      41m     \\ \hline \hline
\multirow{2}{*}{} & \multicolumn{2}{c|}{Printer$^2$}   & \multicolumn{2}{c|}{Trash can$^2$}         & \multicolumn{2}{c|}{\multirow{4}{*}{}} \\ \cline{2-5}
                  & \multicolumn{1}{c|}{SR. $\uparrow$} & ACD. $\downarrow$ & \multicolumn{1}{c|}{SR. $\uparrow$}   & ACD. $\downarrow$   & \multicolumn{2}{c|}{}                  \\ \cline{1-5}
Ours              & \multicolumn{1}{c|}{1/3}   &    76m    & \multicolumn{1}{c|}{2/3}     &     17m     & \multicolumn{2}{c|}{}                  \\ \cline{1-5}
VLMnav            & \multicolumn{1}{c|}{0/3}   &    --    & \multicolumn{1}{c|}{1/3}     &     35m     & \multicolumn{2}{c|}{}                  \\ \hline
\end{tabular}
\end{center}
\end{table}

Fig. \ref{fig:real-world_exp} illustrates key frames from one of the tasks, where the agent starts from the laboratory space and gradually searches for the water dispenser. In the 4th frame, the proposed method chooses to walk towards the door because, through the memory manager, the agent determines that the current site is a laboratory space and therefore suggests leaving the site to find the water dispenser. Conversely, the baseline method demonstrates a propensity to navigate towards the faucet, which leads it into a loop and ultimately results in a failure to locate the target successfully.

\section{Conclusion}

In this work, we propose DyNaVLM, a vision-language navigation framework that leverages self-refining graph memory to enhance VLM-based decision-making. By dynamically encoding topological spatial relations, our approach enables agents to navigate flexibly without relying on fixed action spaces or task-specific training. Our experiments on ObjectNav and GOAT-Bench demonstrate that DyNaVLM surpasses prior VLM-based methods while achieving competitive performance with specialized navigation systems. The results highlight the advantages of memory-augmented reasoning, particularly in long-horizon tasks with complex goal specifications.  

To further validate DyNaVLM’s effectiveness, we conducted real-world evaluations where the agent was tasked with locating objects in diverse environments. The experimental results confirm that DyNaVLM significantly outperforms the baseline in both success rate and efficiency, particularly in multi-target tasks requiring sequential goal identification. The agent demonstrated robust generalization across different starting locations and maintained reliable performance under real-world constraints such as dynamic obstacles and environmental variations. These findings underscore the practical viability of our approach for embodied AI applications beyond simulation-based benchmarks.

Despite these advancements, there are areas for future improvement. The performance gap with specialized methods indicates that integrating additional spatial priors or structured representations may further enhance navigation efficiency. Additionally, while DyNaVLM operates effectively in zero-shot settings, incorporating few-shot adaptation techniques could refine its decision-making over time. As VLMs continue to evolve, we anticipate that frameworks like DyNaVLM will play a pivotal role in bridging the gap between discrete VLN tasks and real-world embodied navigation. Future research should explore real-time learning mechanisms and multi-agent collaboration to further enhance the robustness of these systems, paving the way for more intelligent, interactive, and autonomous robotic assistants.

\addtolength{\textheight}{-12cm}   








\end{document}